\PassOptionsToPackage{table,dvipsnames}{xcolor} 

\documentclass[letterpaper, 10 pt, conference]{ieeeconf}  %

\IEEEoverridecommandlockouts                              %

\usepackage[utf8]{inputenc}
\usepackage[T1]{fontenc}
\usepackage{graphicx}
\usepackage{subcaption} %

\usepackage{amsmath}

\usepackage{listings}
\usepackage{multirow}
\usepackage[final]{changes}
\usepackage{tabularray}
\usepackage[normalem]{ulem}
\usepackage{siunitx}
\usepackage{gensymb}
\usepackage{tablefootnote}
\usepackage{url}
\usepackage{array}
\usepackage{hyperref}
\hypersetup{
    urlcolor=cyan,
    pdftitle={Overleaf Example},
    pdfpagemode=FullScreen,
    }
\usepackage{graphicx}
\usepackage{amssymb}
\usepackage{booktabs}
\usepackage{pifont}
\usepackage{xcolor}

\usepackage{lipsum}  

\definecolor{darkmagenta}{rgb}{0.8, 0.0, 0.8}
\definecolor{darkblue}{rgb}{0.0, 0.0, 0.3}

\newcommand{\xmark}{\ding{55}}%
\newcommand{\mycheckmark}{{\color{darkblue}\checkmark}}
\newcommand{\myxmark}{{\color{darkmagenta}\xmark}}
\newcommand{\myparagraph}[1]{\vspace{3pt}\noindent{\bf #1}}

\newcommand{\ours}{{\cellcolor[gray]{.9}}}

\title{\LARGE \bf
You Only Pose Once: A Minimalist's Detection Transformer\\for Monocular RGB Category-level 9D Multi-Object Pose Estimation
}

\author{{Hakjin Lee}, {Junghoon Seo}, {Jaehoon Sim}
\thanks{PIT IN Co., 
Anyang-si, Gyeonggi-do, Republic of Korea. Email: {\tt\small \{hj, sjh, simjeh\}@pitin-ev.com}}}

\begin{document}

\maketitle
\thispagestyle{empty}
\pagestyle{empty}

 \begin{abstract}

Accurately recovering the full 9-DoF pose of unseen instances within specific categories from a single RGB image remains a core challenge for robotics and automation. Most existing solutions still rely on pseudo-depth, CAD models, or multi-stage cascades that separate 2D detection from pose estimation. Motivated by the need for a simpler, RGB-only alternative that learns directly at the category level, we revisit a longstanding question: \textit{Can object detection and 9-DoF pose estimation be unified with high performance, without any additional data?}
We show that they can be achieved with our method, YOPO, a single-stage, query-based framework that treats category-level 9-DoF estimation as a natural extension of 2D detection. YOPO augments a transformer detector with a lightweight pose head, a bounding-box–conditioned translation module, and a 6D-aware Hungarian matching cost. The model is trained end-to-end only with RGB images and category-level pose labels.
Despite its minimalist design, YOPO sets a new state of the art on three benchmarks. On the REAL275 dataset, it achieves 79.6\% IoU$_{50}$ and 54.1\% under the $10^\circ$$10{\rm{cm}}$ metric, surpassing all prior RGB-only methods and closing much of the gap to RGB-D systems. The code, models, and additional qualitative results can be found on our project page\footnote{\url{https://mikigom.github.io/YOPO-project-page/}}.
 \end{abstract}

\section{Introduction}
 
 The ability to determine an object's three-dimensional position and orientation, known as 6D pose estimation, is a cornerstone of robotic intelligence, enabling critical applications in robotic manipulation~\cite{jian2025monodiff9d,fu20226d,liu2024domain}, augmented reality~\cite{tang20193d,marchand2015pose}, and autonomous driving~\cite{yuan2021temporal,chen2017multi}. While early research focused on estimating the pose of specific, known object instances~\cite{peng2019pvnet,wang2019densefusion,he2020pvn3d}, the practical need to handle novel objects has driven a shift towards category-level pose estimation~\cite{wang2019normalized,chen2021fs,liu2024mh6d}. This more challenging task aims to generalize to previously unseen objects within a given category rather than to only those seen during training.
 
 This work focuses on the particularly challenging and practical setting of \textbf{monocular RGB, category-level, multi-object pose estimation}. Relying on a single RGB image makes this approach highly accessible and cost-effective compared with methods that require active depth sensors~\cite{yang2025rgb, fan2022object}. However, the lack of explicit 3D information introduces significant ambiguity in both object depth and scale. Consequently, the task expands to estimating a 9-Degree-of-Freedom (9-DoF) pose, comprising not only the 3D rotation $R \in SO(3)$ and 3D translation $t \in \mathbb{R}^3$ but also the object's metric 3D size $s \in \mathbb{R}_{> 0}^{3}$ to account for intra-class shape variations~\cite{zhang2024lapose, lee2021category}.
 The ultimate goal of this task is to develop a model and pipeline that, given an RGB image $I_i \in \mathbb{R}^{H \times W \times 3}$, can detect and estimate the pose of all objects present, outputting a set of predictions $\{c_j, R_j, t_j, s_j\}_{j=1}^{M_i}$, where $c_j$ is the object class and $M_i$ is the number of observed objects.

\begin{table}[t!]
\centering
\begin{tabular}{ >{\centering\arraybackslash}m{1.0cm} >{\centering\arraybackslash}m{0.11\textwidth} >{\centering\arraybackslash}m{0.11\textwidth} >{\centering\arraybackslash}m{0.11\textwidth} }
\toprule
 & \textbf{CAD Model?} & \textbf{Instance Mask?} & \textbf{Pseudo-depth?} \\
\midrule
 & \includegraphics[width=\linewidth]{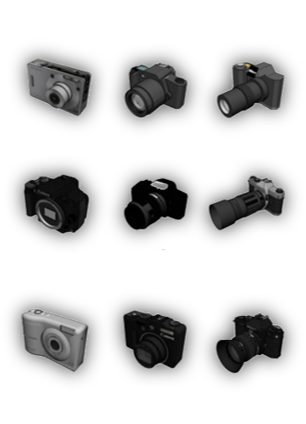} & \includegraphics[width=\linewidth]{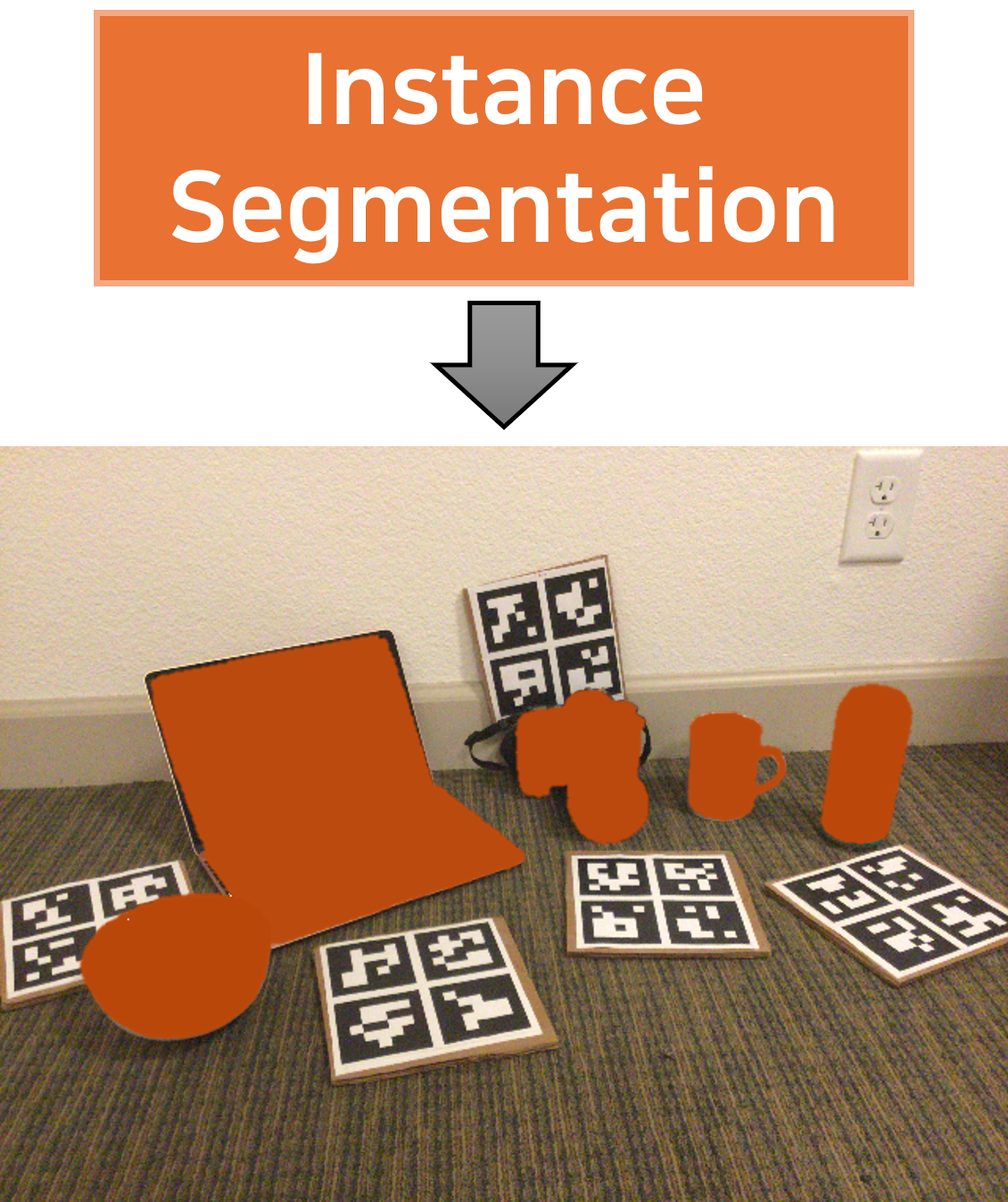} & \includegraphics[width=\linewidth]{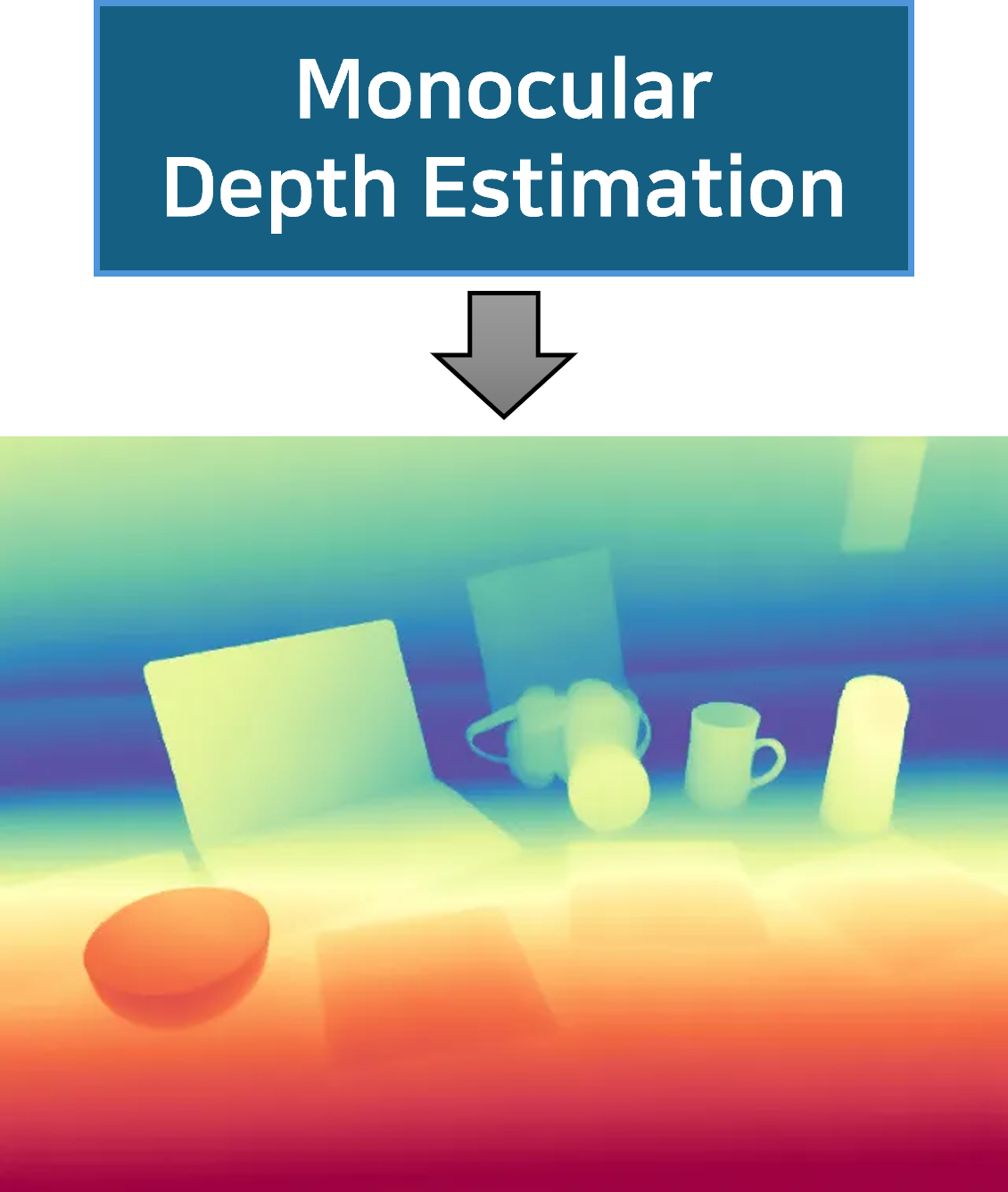} \\
\midrule 
\textbf{Others} & ~\cite{fan2022object,wei2024rgb,zhang2024lapose} & ~\cite{huang2025givepose,lee2021category,chen2020category} & ~\cite{yang2025rgb,jian2025monodiff9d,wei2024rgb} \\
\addlinespace
\textbf{Ours} & \myxmark & \myxmark & \myxmark \\
\bottomrule 
\end{tabular}
\end{table}
\begin{figure}[t!]
    \centering
    \includegraphics[width=0.99\linewidth]{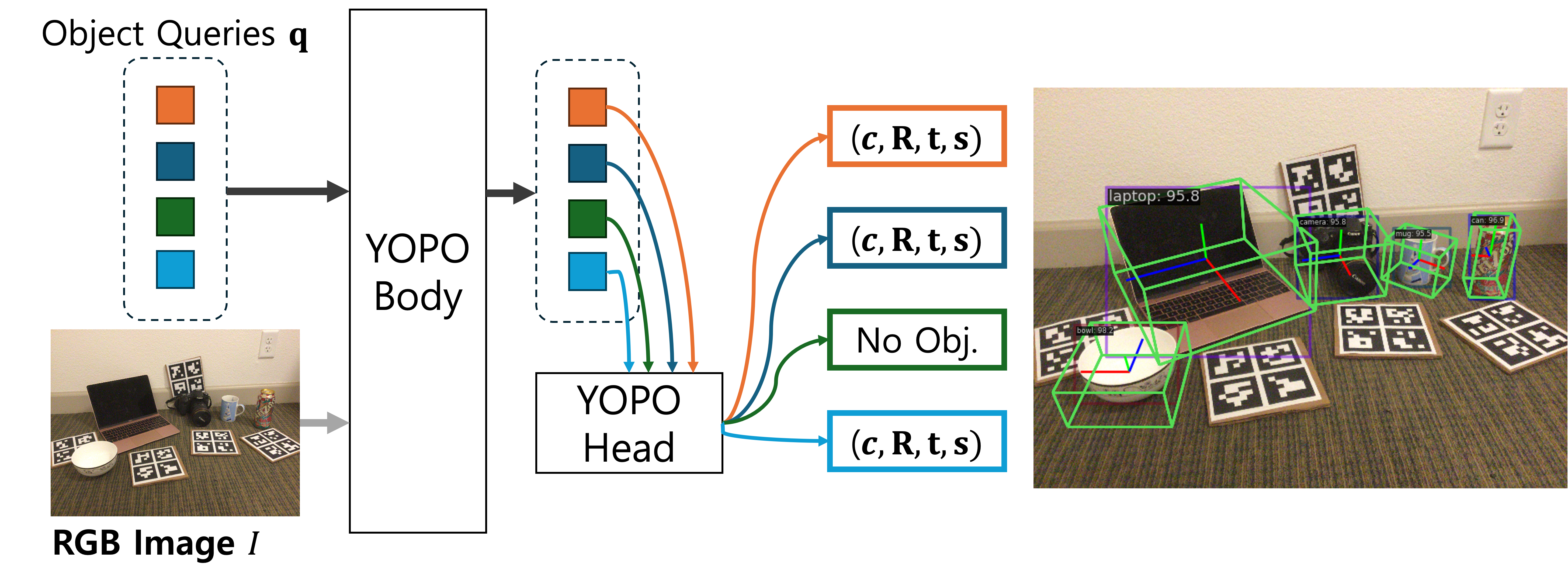}
\caption{\textbf{Main contribution of this paper.} Unlike prevailing category-level pose estimation methods that rely on external geometric priors such as 3D CAD models, instance segmentation masks, or pseudo-depth maps (top), our framework is end-to-end and requires none of these (bottom). Using only a raw RGB image as input, YOPO delivers state-of-the-art joint detection and 9D pose estimation for all objects in a single forward pass, with no intermediate steps or post-processing.}
    \label{fig:enter-label}
    \vspace{-5mm}
\end{figure}

Despite significant progress, most leading approaches are not truly end-to-end. Instead, they rely on complex, multistage pipelines. Furthermore, these methods often require auxiliary data. The data usually include category-specific shape priors from 3D CAD models~\cite{fan2022object,chen2021sgpa,huang2025givepose}, instance segmentation masks for initial object localization~\cite{yang2025rgb,jian2025monodiff9d,zhang2024lapose}, or estimated pseudo-depth maps to simplify 3D reasoning~\cite{yang2025rgb,wei2024rgb}. Such dependencies hinder end-to-end training, increase computational overhead, and create performance bottlenecks that depend on these external modules.

This paper questions the necessity of such complex pipelines. We draw inspiration from the success of the modern query-based detection transformer (DETR)~\cite{carion2020end, zhang2023dino}, which has demonstrated the power of formulating detection as a direct set prediction problem. We investigate whether this end-to-end paradigm can be extended from 2D detection to the challenging 9-DoF 3D pose estimation domain. To this end, we introduce \textbf{YOPO}: a new framework for monocular, category-level 9D pose estimation that operates in a truly end-to-end fashion. YOPO is built upon a transformer-based object detector and learns to directly predict an object's bounding box, class, 3D rotation, 3D translation, and 3D scale in a single forward pass. It is trained jointly using only raw RGB images and their corresponding category-level 9D pose annotations. Crucially, YOPO dispenses with the need for 3D CAD models, shape priors, instance segmentation masks, pseudo-depth maps, or even explicit 2D bounding boxes during the training and inference stages. Our experiments show that this simpler, more direct approach not only streamlines the process but also establishes a new state of the art on standard benchmarks.
 
As shown in Fig.~\ref{fig:enter-label}, our main contributions are as follows:
\begin{itemize}
  \item We propose YOPO, a novel single-stage, query-based framework for monocular category-level 9D object pose estimation that is fully end-to-end trainable and requires only RGB images and 9D pose labels.
  \item We present a minimalist yet effective design that augments a detection transformer with bounding box–conditioned 2D center and depth regression for stable 3D translation recovery, and a 6D-aware bipartite matching cost.
  \item Through extensive experiments on the REAL275, CAMERA25, and HouseCat6D benchmarks, we demonstrate that YOPO significantly outperforms previous, more complex methods and sets a new state of the art in monocular category-level pose estimation.
\end{itemize}

\section{Related Work}

\subsection{Data Requirements for RGB 9D Object Pose Estimation}
A closer inspection of existing research in monocular RGB category-level pose estimation reveals that few methods rely strictly on RGB images and their corresponding 9D pose annotations for both training and inference. Many state-of-the-art pipelines incorporate additional data and assumptions to achieve high performance.

A common requirement is the use of 3D CAD models of objects within the training categories, even if they are not required at inference time. These models are often used either to construct a canonical representation (e.g., NOCS~\cite{wang2019normalized})~\cite{lee2021category,huang2025givepose,zhang2024lapose} or to generate category-level shape priors that guide the learning process~\cite{fan2022object,wei2024rgb}. Another prevalent dependency is the use of instance segmentation masks. Methods such as MSOS~\cite{lee2021category}, DMSR~\cite{wei2024rgb}, LaPose~\cite{zhang2024lapose}, MonoDiff9D~\cite{jian2025monodiff9d}, and DA-Pose~\cite{yang2025rgb} typically employ an off-the-shelf instance segmentation model (e.g., Mask R-CNN~\cite{he2017mask}) to isolate objects from the background. These masks are essential for cropping input images or feature maps to the object's region of interest. Some methods incorporate pseudo-depth information by leveraging pretrained monocular depth estimators (e.g., ZoeDepth~\cite{bhat2023zoedepth} or DepthAnything~\cite{yang2024depth}). This allows them to recover metric depth~\cite{lee2021category,fan2022object,yang2025rgb} or relative depth~\cite{wei2024rgb}, effectively converting the monocular problem into a pseudo-RGB-D one to simplify 3D reasoning.

In contrast, our approach operates without any of these additional data dependencies. To the best of our knowledge, the only notable prior work with similarly minimal data assumptions is CenterPose~\cite{lin2022single}. However, the performance of this keypoint-based approach has been surpassed by the aforementioned more complex pipelines that leverage additional data~\cite{yu2023cattrack,mei2025multi}. This highlights a gap in the literature for a method that can achieve state-of-the-art performance while adhering to a strict monocular RGB formulation. Our work aims to fill this gap, demonstrating that it is possible to surpass the performance of recent methods without resorting to external data sources such as CAD models, segmentation masks, or pseudo-depth maps (as shown in Table~\ref{table:method_details} and Table~\ref{table:benchmark_comparison}).

\begin{table}[t!]
\renewcommand\arraystretch{1.1}
\centering
\vspace{0.7em}
\caption{Comparison of methods in terms of additional data requirements. \mycheckmark{} denotes required, and \myxmark{} denotes not required. We compare the RGB-only versions of \emph{Synthesis}~\cite{chen2020category} and \emph{CenterSnap}~\cite{irshad2022centersnap}.}
\vspace{-0.7em}
\setlength{\tabcolsep}{4.6pt}{
    \begin{tabular}{c|cccc}
    \hline
    Method & CAD Model & Seg. Mask & Psudo-depth \\
    \hline
    Synthesis (ECCV `20) \cite{chen2020category} & \mycheckmark & \mycheckmark & \myxmark \\
    MSOS (RA-L `21) \cite{lee2021category} & \mycheckmark & \mycheckmark & \myxmark \\
    CenterSnap (ICRA `22) \cite{irshad2022centersnap} & \mycheckmark & \myxmark & \myxmark  \\
    OLD-Net (ECCV `22) \cite{fan2022object} & \mycheckmark & \mycheckmark & \mycheckmark \\
    DMSR (ICRA `24) \cite{wei2024rgb} & \mycheckmark & \mycheckmark & \mycheckmark  \\
    LaPose (ECCV `24) \cite{zhang2024lapose} & \mycheckmark & \mycheckmark & \myxmark  \\
    MonoDiff9D (ICRA `25) \cite{jian2025monodiff9d} & \myxmark & \mycheckmark & \mycheckmark  \\
    DA-Pose (RA-L `25) \cite{yang2025rgb} & \myxmark & \mycheckmark & \mycheckmark \\
    GIVEPose (CVPR `25) ~\cite{huang2025givepose} & \mycheckmark & \mycheckmark & \myxmark  \\ \midrule
    \textbf{YOPO (Ours)} & \myxmark & \myxmark & \myxmark  \\
    \bottomrule
    \end{tabular}}
\label{table:method_details}
\vspace{-5mm}
\end{table}

\subsection{Query-based Oriented Object Detection}
Our research is motivated by the significant success of oriented object detection~\cite{xie2021oriented,ding2019learning} in the broader computer vision field. This 2D task aims to extract each object's class $c$, 2D location $t \in \mathbb{R}^2$, in-plane rotation $R \in SO(2)$, and size $s \in \mathbb{R}_{>0}^{2}$ from a single RGB image. Recently, query-based architectures, pioneered by DETR~\cite{carion2020end,zhang2023dino}, have shown strong competitiveness by formulating the task as a direct set prediction problem~\cite{zeng2024ars,lee2025hausdorff}.

A key advantage of these query-based 2D detectors is their ability to operate in a clean, end-to-end fashion, requiring no additional data or priors beyond the input image and corresponding annotations. This contrasts with the complex pipelines commonly seen in 9D pose estimation. From this perspective, a central question motivating our work is whether the success of this streamlined, query-based paradigm can be transferred from 2D object detection to the more challenging domain of RGB, category-level 9D pose estimation. Notably, while such minimalist end-to-end detection approaches are more common in related domains like RGB-D~\cite{irshad2022centersnap,irshad2022shapo} or model-level~\cite{maji2024yolo,jantos2023poet,tekin2018real,xiang2018posecnn} pose estimation, they remain particularly scarce in the challenging RGB-only, category-level setting.

\section{Method}

\begin{figure*}[t]
  \centering
  \vspace{0.3em}
  \begin{subfigure}[c]{0.55\textwidth}
    \centering
    \includegraphics[width=\textwidth]{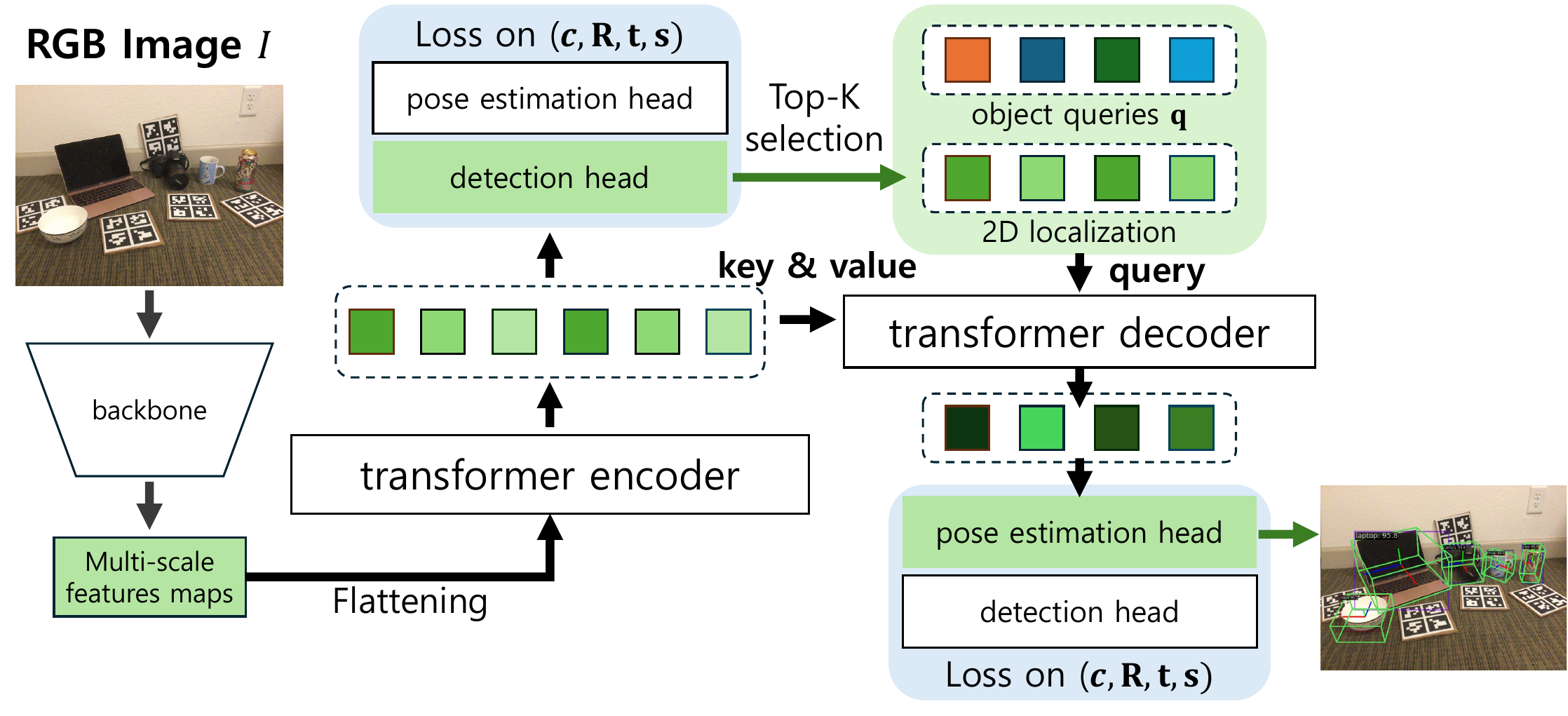}
    \caption{Architecture overview.}
    \label{fig:architecture_a}
  \end{subfigure}
  \hfill
  \begin{subfigure}[c]{0.38\textwidth}
    \centering
    \vspace{5mm}
    \includegraphics[width=\textwidth]{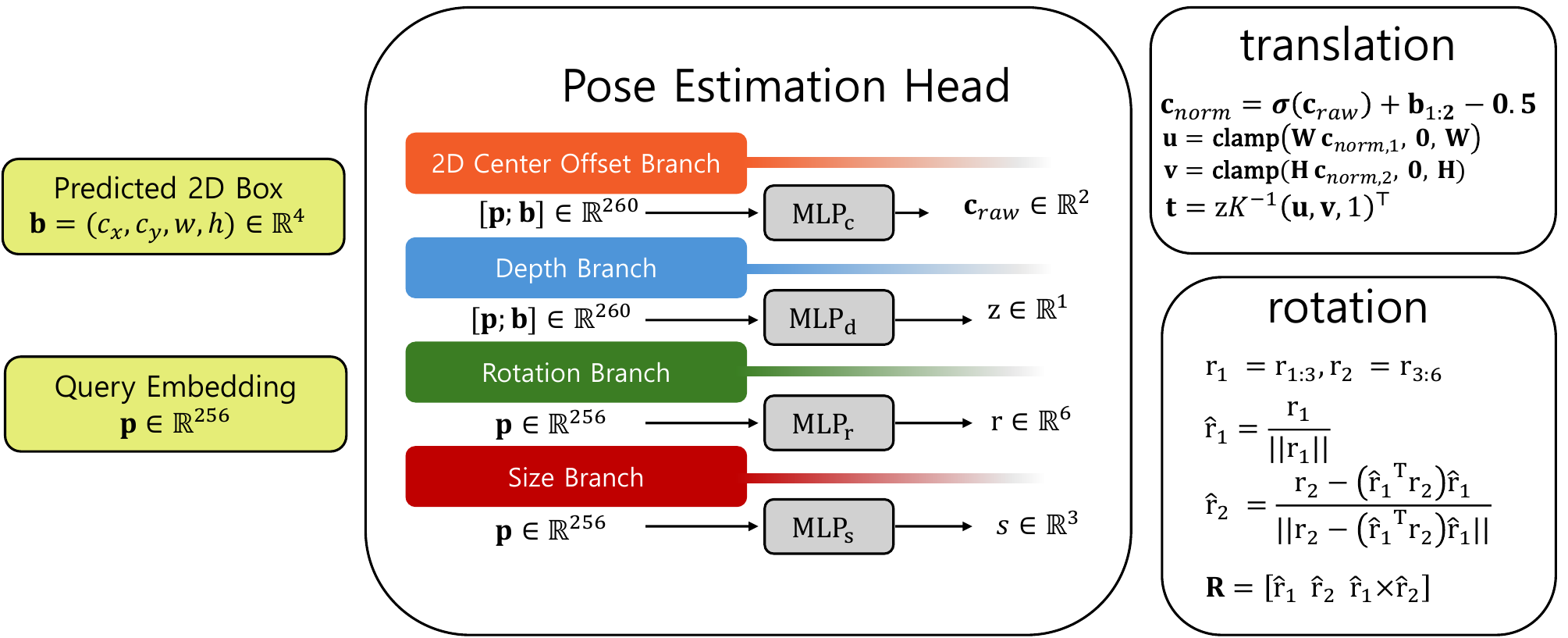}
    \vspace{3mm}
    \caption{Bounding box-conditioned 3D prediction.}
    \label{fig:architecture_b}
  \end{subfigure}
  \caption{
Overview of our method. (a) The model predicts object properties from transformer-decoder outputs using task-specific heads. (b) The translation and depth head estimates 2D center locations as offsets from bounding-box centers, enabling 3D translation and depth recovery via back-projection. Predicted bounding boxes are concatenated with the input query to provide spatial information for accurate 2D center and depth estimation.
  }
  \label{fig:architecture}
  \vspace{-3mm}
\end{figure*}

\subsection{Overall Architecture}
\label{subsec:overview}

Given an input RGB image $I \in \mathbb{R}^{H \times W \times 3}$ and the camera intrinsic matrix $K \in \mathbb{R}^{3 \times 3}$, our goal is to estimate, in a single forward pass, the category and 9D pose of all object instances:
\begin{equation}
\hat{y}_i = \left( c_i, \mathbf{R}_i, \mathbf{t}_i, \mathbf{s}_i \right), \quad i = 1, \dots, N,
\label{eq:pose_output}
\end{equation}
where $c_i$ denotes the object category, $\mathbf{R}_i \in \mathrm{SO}(3)$ the 3D rotation, $\mathbf{t}_i \in \mathbb{R}^3$ the translation, and $\mathbf{s}_i \in \mathbb{R}^3$ the anisotropic scale.

Existing methods~\cite{jian2025monodiff9d, wei2024rgb, huang2025givepose} typically decompose this task into two stages. In the first stage, they predict 2D object detections $(c_i, \mathbf{b}_i)$, where $\mathbf{b}_i \in \mathbb{R}^4$ denotes the 2D bounding box parameterized by $(x, y, w, h)$. In the second stage, pose estimation is performed on cropped image regions that contain individual object instances.
To isolate objects from the background, these methods often employ separately trained instance segmentation models. Moreover, during training, collections of CAD models are commonly used to provide shape priors, enabling the pose estimation network to leverage existing geometric information. While effective, these dependencies on segmentation masks and CAD priors increase annotation costs and hinder generalization to novel object categories.

In contrast, our approach follows the design principles of DETR, which eliminate handcrafted priors. Our model directly predicts $(c_i, \mathbf{R}_i, \mathbf{t}_i, \mathbf{s}_i)$ from RGB images, end to end, using only object-category, pose, and size annotations. This design simplifies training and improves scalability across diverse object categories, without requiring instance masks or CAD models.

Specifically, our model builds on the transformer-based detector DINO~\cite{zhang2023dino}, which extends DETR~\cite{carion2020end} with a two-stage refinement mechanism. As illustrated in Fig.~\ref{fig:architecture_a}, the architecture comprises:
(1) a multi-scale feature backbone for image feature extraction,
(2) a transformer encoder that processes the feature maps,
(3) a transformer decoder that refines object queries, and
(4) task-specific prediction heads applied at both the proposal and refinement stages.

In the first stage, the transformer encoder takes the flattened multi-scale backbone features as input and outputs memory features enriched with global context. A detection head then predicts a set of reference points with associated objectness scores. The top-K proposals are selected based on these scores and are used as spatial anchors (2D reference points) with learnable object queries $\mathbf{q}$. In the second stage, the transformer decoder takes these queries and iteratively updates them through cross-attention with the encoder's memory features to produce enriched embeddings $\mathbf{p}$. These refined queries are subsequently fed into parallel prediction heads to explicitly derive the 9-DoF parameters: rotation ($\mathbf{R}_i$), translation ($\mathbf{t}_i$), and scale ($\mathbf{s}_i$).

\subsection{Parallel Prediction Heads}
\label{subsec:heads}

At both the proposal and refinement stages, our architecture employs two parallel heads: a detection head and a pose-estimation head.

\myparagraph{Detection Head.}
The detection head predicts object categories $c_i$ and 2D bounding boxes $\mathbf{b}_i$. In the proposal stage, it generates coarse localization cues and selects the top-K object queries $\mathbf{q}$. During the refinement stage, it provides auxiliary supervision on categories and boxes; however, its predictions are not used at inference.

Although we optimize the detection head with a 2D bounding-box loss, these boxes require no manual annotation, as they can be automatically derived by projecting the annotated 3D cuboids onto the image plane (Table~\ref{tab:ablation_projbox_comparison}).

\myparagraph{Pose Estimation Head.}
The pose estimation head takes the object queries and predicts the 9-DoF parameters $(\mathbf{R}_i, \mathbf{t}_i, \mathbf{s}_i)$ via four specialized multi-layer perceptron (MLP) branches: (1) 2D center offset, (2) depth, (3) rotation, and (4) scale. Specifically, the scale head directly regresses the 3D anisotropic scale $\mathbf{s}_i$. By leveraging the global context encoded in the query, it predicts absolute scale without relying on categorical shape priors. To alleviate monocular ambiguity, both the center and depth heads predict their values by explicitly conditioning the query on the 2D bounding box (detailed in the following subsection). Rotation is predicted via a continuous 6D representation.
The 3D translation $\mathbf{t}_i$ is reconstructed by back-projecting the predicted 2D center and depth using the camera intrinsics $K$. This head is supervised at both the proposal and refinement stages, ensuring queries are geometry-aware early on.

By sharing object queries, our parallel architecture jointly optimizes 2D localization and 3D reasoning, enabling the two tasks to mutually reinforce each other.

\subsection{2D Bounding Box-Conditioned 3D Prediction}
For 3D translation, we employ a disentangled parameterization following Simonelli et al.~\cite{simonelli2019disentangling}, wherein the image-plane projected center $(u_i, v_i)$ and physical depth $z_i$ are predicted separately. This decoupled approach significantly enhances training stability in monocular settings.

\myparagraph{Bounding Box-Conditioned Center Prediction.}
Following the disentangled translation formulation of Simonelli et al.~\cite{simonelli2019disentangling},
we predict the image-plane center as an offset from the center of the predicted 2D bounding box.
Let $\mathbf{p}_i \in \mathbb{R}^D$ denote the refined embedding of the $i$-th object from the transformer decoder,
and $\mathbf{b}_i \in \mathbb{R}^4$ the predicted bounding box parameterized as
$\mathbf{b}_i=(c_{x,i},c_{y,i},w_i,h_i)$.
The first two components $\mathbf{b}_{i,1:2}\in\mathbb{R}^2$ correspond to the box center.

Instead of regressing the center solely from the object query,
we condition the prediction on the bounding box by concatenating $\mathbf{p}_i$ and $\mathbf{b}_i$:
\begin{equation}
\mathbf{c}_{raw,i}
=
\operatorname{MLP}_c
\big(
\operatorname{Concat}(\mathbf{p}_i,\mathbf{b}_i)
\big),
\quad
\operatorname{MLP}_c:\mathbb{R}^{D+4}\rightarrow\mathbb{R}^2.
\label{eq:c_raw}
\end{equation}

The predicted residual is combined with the box center and converted to a normalized center coordinate via
\begin{equation}
\mathbf{c}_{norm,i}
=
\sigma(\mathbf{c}_{raw,i})+\mathbf{b}_{i,1:2}
-
\begin{bmatrix}
0.5\\
0.5
\end{bmatrix},
\quad
\mathbf{c}_{norm,i}\in\mathbb{R}^2,
\label{eq:c_norm}
\end{equation}
where $\sigma(\cdot)$ denotes the element-wise sigmoid.
Crucially, the model is supervised directly on these normalized coordinates $\mathbf{c}_{norm,i} \in [0, 1]^2$ during training, which ensures stability across varying image resolutions.

During inference, the normalized center is scaled by the image width $W$ and height $H$ to recover the absolute pixel coordinates:
\begin{equation}
\begin{split}
u_i&=\mathrm{clamp}(W \cdot \mathbf{c}_{norm,i,1}, 0, W),\\
v_i&=\mathrm{clamp}(H \cdot \mathbf{c}_{norm,i,2}, 0, H).
\end{split}
\label{eq:uv}
\end{equation}

Finally, the 3D translation is recovered by perspective back-projection using the predicted depth $z_i$:
\begin{equation}
\mathbf{t}_i
=
z_i K^{-1}
\begin{bmatrix}
u_i\\
v_i\\
1
\end{bmatrix}.
\label{eq:backprojection}
\end{equation}

Conditioning the center prediction on the bounding box provides explicit geometric guidance,
allowing the model to exploit spatial cues encoded in the detection head while preserving the
structural benefits of the disentangled translation formulation.

\myparagraph{Bounding Box-Conditioned Depth Prediction.}
Similar to the center prediction head, we condition the depth predictor by concatenating the object query with the 2D bounding-box parameters. This explicit geometric conditioning provides complementary spatial cues, which stabilizes depth regression, reduces scale ambiguities, and ultimately improves 3D translation accuracy.

\subsection{3D Matching Costs for Bipartite Matching}

We follow DETR~\cite{carion2020end} and use one-to-one bipartite matching to assign predictions to ground-truth instances. The base matching cost includes classification, 2D bounding-box regression, and intersection-over-union (IoU) terms:
\begin{equation}
\mathcal{C}_{\text{match}} =
  \lambda_{\text{cls}}^{\mathcal{C}} \cdot \mathcal{C}_{\text{cls}} +
  \lambda_{\text{bbox}}^{\mathcal{C}} \cdot \mathcal{C}_{\text{bbox}} +
  \lambda_{\text{IoU}}^{\mathcal{C}} \cdot \mathcal{C}_{\text{IoU}},
\label{eq:match_cost_base}
\end{equation}
where $\mathcal{C}_{\text{cls}}$ is the classification cost, $\mathcal{C}_{\text{bbox}}$ is the L1 distance between the predicted and ground-truth 2D bounding boxes, $\mathcal{C}_{\text{IoU}}$ is the negative IoU, and $\lambda^{\mathcal{C}}$ denotes the corresponding weights.

While standard DETR relies solely on 2D proximity, we explicitly regularize 3D structure by adding translation and rotation terms: $\mathcal{C}_{\text{trans}}$ is the Euclidean distance between 3D translations, and $\mathcal{C}_{\text{rot}}$ is the geodesic distance between rotation matrices (accounting for object symmetries). The final 6D-aware matching cost is
\begin{align}
\mathcal{C}_{\text{match}} =\;
& \lambda_{\text{cls}}^\mathcal{C}   \cdot \mathcal{C}_{\text{cls}}   +
  \lambda_{\text{bbox}}^\mathcal{C}  \cdot \mathcal{C}_{\text{bbox}}  +
  \lambda_{\text{IoU}}^\mathcal{C}   \cdot \mathcal{C}_{\text{IoU}} \notag \\
& + \lambda_{\text{trans}}^\mathcal{C} \cdot \mathcal{C}_{\text{trans}} +
    \lambda_{\text{rot}}^\mathcal{C}   \cdot \mathcal{C}_{\text{rot}}.
\label{eq:matching_cost}
\end{align}
We use the following default weights:
$\lambda^{\mathcal{C}}_{\text{cls}}{=}2.0$,
$\lambda^{\mathcal{C}}_{\text{bbox}}{=}5.0$,
$\lambda^{\mathcal{C}}_{\text{IoU}}{=}2.0$,
$\lambda^{\mathcal{C}}_{\text{trans}}{=}5.0$,
$\lambda^{\mathcal{C}}_{\text{rot}}{=}2.0$.
We omit 3D scale from $\mathcal{C}_{\text{match}}$. The base cost ($\mathcal{C}_{\text{bbox}}$ and $\mathcal{C}_{\text{IoU}}$) already provides a strong proxy for apparent object size. Furthermore, explicit 3D scale prediction from a monocular image is inherently ambiguous, especially early in training. Including it in the cost matrix can introduce noisy query assignments. Thus, we decouple scale from the matching process and optimize it purely via the loss function post-assignment.

\subsection{Implementation Details}

\myparagraph{Architecture.}
Our framework is built on top of the transformer-based DINO detector~\cite{zhang2023dino}. We largely follow DINO’s default settings, including the multi-scale backbone, encoder-decoder transformer structure, and a two-stage refinement with denoising training. Specifically, we adopt the same number of encoder and decoder layers, attention heads, hidden dimensions, and feed-forward network settings as DINO. Each task-specific component in the pose estimation head also follows the MLP structure of the detection head, except for the output dimension (or the input dimension when bounding-box information is concatenated with the object queries). Unlike DINO, we reduce the number of object queries to 100.

\myparagraph{Pose Estimation Head.}
We represent rotations using the continuous 6D parameterization~\cite{zhou2019continuity}, mapped to $\mathrm{SO}(3)$ via Gram--Schmidt-like orthonormalization, and supervise them with a geodesic loss. Both the depth $z_i$ and the anisotropic 3D scale $\mathbf{s}_i \in \mathbb{R}^3$ are regressed directly in linear space. Unless otherwise stated, we adopt class-wise (\textbf{CW}) rotation and scale prediction heads: the output dimensions are expanded by the number of categories, and the final predictions are selected according to the highest class confidence.

\myparagraph{Losses and Training.}
We jointly optimize detection and pose using
\begin{align}
\mathcal{L}_{\text{total}} = \;
& \lambda_{\text{cls}} \mathcal{L}_{\text{cls}} +
\lambda_{\text{bbox}} \mathcal{L}_{\text{bbox}} +
\lambda_{\text{IoU}} \mathcal{L}_{\text{IoU}} \nonumber \\ &+
\lambda_{\text{center2D}} \mathcal{L}_{\text{center2D}} +
\lambda_{\text{depth}} \mathcal{L}_{\text{depth}} \nonumber
\\ &+\lambda_{\text{rot}} \mathcal{L}_{\text{rot}} +
\lambda_{\text{scale}} \mathcal{L}_{\text{scale}}.
\end{align}
We use focal loss~\cite{lin2017focal} for classification $\mathcal{L}_{\text{cls}}$, L1 for box $\mathcal{L}_{\text{bbox}}$ and center prediction $\mathcal{L}_{\text{center2D}}$, GIoU~\cite{rezatofighi2019generalized} for IoU loss $\mathcal{L}_{\text{IoU}}$, and L2 for depth $\mathcal{L}_{\text{depth}}$ and scale $\mathcal{L}_{\text{scale}}$. Based on our ablation experiments, we adopt the following optimal weight configuration for the pose terms: $\lambda_{\text{rot}}=5.0$, $\lambda_{\text{center2D}}=5.0$, $\lambda_{\text{depth}}=50.0$, and $\lambda_{\text{scale}}=5.0$, alongside standard DETR weights ($\lambda_{\text{cls}}=1.0$, $\lambda_{\text{bbox}}=5.0$, $\lambda_{\text{IoU}}=2.0$).
We use AdamW~\cite{loshchilov2017decoupled} with learning rate $1{\times}10^{-4}$ for batch size 16 on 4$\times$A6000 GPUs.
All models are initialized from DINO pretrained on COCO to leverage strong object detection performance.

\section{Experiments}

\begin{table*}[t!]
\renewcommand\arraystretch{1.2}
\newcommand{\tabincell}[2]{\begin{tabular}{@{}#1@{}}#2\end{tabular}}
\centering
\vspace{0.5em}
\caption{Comparison on the CAMERA25 and REAL275 datasets. For each result, \textbf{bold} is used to indicate the top-performing method among all methodologies. An \underline{underline} highlights the best performing method \textit{when our proposed method is excluded}.}
\vspace{-0.7em}
\setlength{\tabcolsep}{4.6pt}{
    \begin{tabular}{c|ccccc|ccccc}
    \hline
    \multirow{2}[2]{*}{Method} & \multicolumn{5}{c}{CAMERA25} & \multicolumn{5}{|c}{REAL275}\\
\cline{2-11}    & ${\rm{IoU}_{50}}$ & ${\rm{IoU}_{75}}$ & \tabincell{c}{$10{\rm{cm}}$} & \tabincell{c}{$10^\circ$} & \tabincell{c}{$10^\circ$$10{\rm{cm}}$} & ${\rm{IoU}_{50}}$ & ${\rm{IoU}_{75}}$ & \tabincell{c}{$10{\rm{cm}}$} & \tabincell{c}{$10^\circ$} & \tabincell{c}{$10^\circ$$10{\rm{cm}}$}\\
    \hline
    Synthesis (ECCV 2020) \cite{chen2020category} & - & - & - & - & - & - & - & 34.0 & 14.2 & 4.8 \\
    MSOS (RA-L 2021) \cite{lee2021category} & 32.4 & 5.1 & 29.7 & 60.8 & 19.2 & 23.4 & 3.0 & 39.5 & 29.2 & 9.6\\
    CenterSnap-RGB (ICRA 2022) \cite{irshad2022centersnap} & - & - & - & - & - & 31.5 & - & - & - & 30.1 \\
    OLD-Net (ECCV 2022) \cite{fan2022object} & 32.1 & 5.4 & 30.1 & 74.0 & 23.4 & 25.4 & 1.9 & 38.9 & 37.0 & 9.8\\
    FAP-Net (ICRA 2024) \cite{li2024implicit} & 39.2 & 6.7 & 36.0 & 80.4 & \underline{29.8} & \underline{36.8} & 5.2 & \underline{49.7} & 49.6 & 24.5\\
    DMSR (ICRA 2024) \cite{wei2024rgb} & 34.6 & 6.5 & 32.3 & \underline{81.4} & 27.4 & 28.3 & 6.1 & 37.3 & \underline{59.5} & 23.6\\
    LaPose (ECCV 2024) \cite{zhang2024lapose} & - & - & - & - & - & 17.5 & 2.6 & 44.4 & - & 30.5 \\
    MonoDiff9D (ICRA 2025) \cite{jian2025monodiff9d} & 35.2 & \underline{6.7} & 33.6 & 80.1 & 28.2 & 31.5 & \underline{6.3} & 41.0 & 56.3 & 25.7\\
    DA-Pose (RA-L 2025) \cite{yang2025rgb} & \underline{41.4} & 6.1 & \underline{40.5} & 60.8 & 24.6 & 28.1 & 3.6 & 45.8 & 27.5 & 13.4 \\
    GIVEPose (CVPR 2025) ~\cite{huang2025givepose} & - & - & - & - & - & 20.1 & - & 45.9 & - & \underline{34.2} \\
    \midrule
 YOPO R50 (Ours) & 41.4 & 7.9 & 36.3 & 78.2 & 30.1 & 67.1  & 16.6 & 75.6 & 54.0 & 40.7 \\
 YOPO Swin-L (Ours) & \textbf{46.6}  & \textbf{11.8} & \textbf{43.1} & \textbf{88.7}& \textbf{38.7} & 71.6  & 16.4 & 77.8 & \textbf{69.6} & 52.8 \\
    YOPO Swin-L$^*$ (Ours) & - & - & - & - & -  & \textbf{79.6}  & \textbf{19.6} & \textbf{84.4} & 66.0 & \textbf{54.1}  \\
        \bottomrule
    \end{tabular}}
\label{table:benchmark_comparison}
\end{table*}

\subsection{Datasets and Metrics}
\myparagraph{Datasets.}
We evaluate YOPO on three widely used benchmarks for category-level 9D pose estimation: CAMERA25, REAL275~\cite{wang2019normalized}, and HouseCat6D~\cite{jung2024housecat6d}.
CAMERA25 is a synthetic dataset containing 275K training images and 25K test images across six object categories.
REAL275 is a real-world dataset with the same categories as CAMERA25. It consists of 4.3K training images from 7 scenes and 2.75K test images from 6 scenes, with three unseen object instances per category in the test split.
HouseCat6D features 194 high-fidelity 3D object models across 10 household categories, with 20K training, 1.4K validation, and 3K test images captured in 41 real scenes.

\myparagraph{Evaluation Metrics.}
We follow standard protocols~\cite{wang2019normalized, chen2021fs} and report two main metrics:

\begin{itemize}
  \item \textbf{3D IoU}: We report mean Average Precision (mAP) at 3D bounding-box IoU thresholds of 50$\%$ and 75$\%$. This metric jointly reflects the accuracy of pose and 3D scale estimation.
  \item \textbf{$n^\circ$ $m$\,cm}: We also report the mAP of predictions in which the rotation error is below $n^\circ$ and the translation error is under $m$\,cm. This directly evaluates geometric accuracy and is commonly used for 6D pose estimation.
\end{itemize}
Following DMSR~\cite{wei2024rgb}, we adopt 50$\%$ and 75$\%$ as the thresholds for 3D IoU evaluation, and we report pose accuracy under the $10^\circ$, $10\,\mathrm{cm}$, and $10^\circ\text{-}10\,\mathrm{cm}$ criteria for assessing rotation and translation errors.

\begin{figure}
    \centering
    \vspace{0.5em}
    \includegraphics[width=0.48\textwidth]{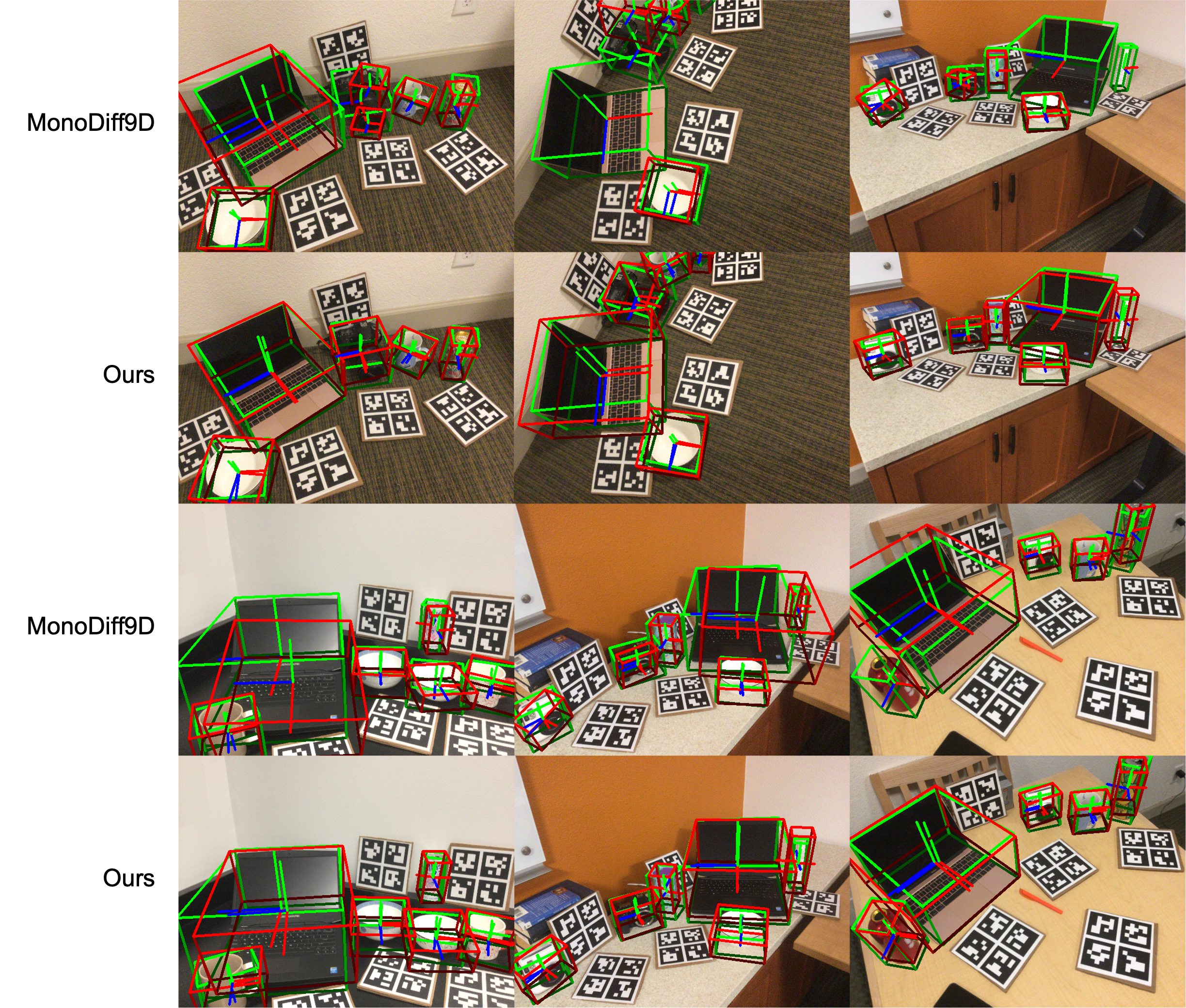}
    \caption{Qualitative comparison of pose estimation results on the REAL275 dataset. We compare our model with MonoDiff9D~\cite{jian2025monodiff9d}. Predicted poses are shown in red, while ground-truth annotations are shown in green.}
    \label{fig:quali_comparison}
    \vspace{-5mm}
\end{figure}

\myparagraph{Experimental Settings.}
Unless otherwise specified, we train YOPO for 12 epochs on the combined CAMERA25 and REAL275 datasets using 9D pose annotations and 2D bounding boxes, and evaluate using a single model. YOPO$^*$ is obtained by fine-tuning YOPO on REAL275 for 12 epochs. For HouseCat6D, we adopt the same training schedule and hyperparameters, but set the learning rate to $2 \times 10^{-4}$, which is twice that used for CAMERA25 and REAL275. We apply random pixel translations and horizontal flips for data augmentation.

\subsection{Comparison with Previous Methods}
\label{subsec:comparison_sota}
Table~\ref{table:benchmark_comparison} compares YOPO with existing state-of-the-art methods on the CAMERA25 and REAL275 datasets.
YOPO consistently outperforms all prior RGB-only approaches on both datasets. On CAMERA25, YOPO with a Swin-L backbone~\cite{liu2021swin} achieves \textbf{46.6}\% $\mathrm{IoU}_{50}$, \textbf{11.8}\% $\mathrm{IoU}_{75}$, and \textbf{38.7}\% under the $10^\circ$$10{\rm{cm}}$ criterion.
On REAL275, YOPO achieves \textbf{71.6}\% $\mathrm{IoU}_{50}$ and \textbf{52.8}\% $10^\circ$$10{\rm{cm}}$, outperforming prior RGB-only methods across all metrics. With additional fine-tuning on the REAL275 training split (denoted as YOPO$^*$), our method further improves to \textbf{79.6}\% $\mathrm{IoU}_{50}$ and \textbf{54.1}\% $10^\circ$$10{\rm{cm}}$. YOPO with the lighter ResNet-50 backbone~\cite{he2016deep} also maintains this overall superiority.

Figure~\ref{fig:quali_comparison} presents qualitative results comparing our method with the previous state-of-the-art MonoDiff9D~\cite{jian2025monodiff9d} on the REAL275 dataset.
The top two rows illustrate the advantages of our end-to-end framework for joint object detection and pose estimation. Unlike MonoDiff9D, which relies on a separately trained instance segmentation model and suffers from missed detections and false positives, YOPO directly detects objects and estimates their poses in a unified pipeline, thereby reducing error propagation.
The bottom two rows demonstrate the accuracy of our method in estimating 3D translation, rotation, and scale. YOPO's predictions consistently align more closely with the ground-truth cuboids than those of MonoDiff9D, especially in cluttered scenes with varying object scales.

\begin{table}[t]
\renewcommand{\arraystretch}{1.05}
\setlength{\tabcolsep}{3.8pt}
\centering
\caption{Ablation on \textbf{REAL275}.
\mycheckmark: component enabled, \myxmark: disabled. The shaded rows indicate the selection for our final model, corresponding to the main table (Table~\ref{table:benchmark_comparison}).
}
\label{tab:ablation_real275_main}
\vspace{.3em}

\resizebox{0.5\textwidth}{!}{%
\begin{tabular}{lccccc|cccccc}
\toprule
 & BC & Aug & CW & WS & RFT &
${\rm{IoU}_{50}}$ & ${\rm{IoU}_{75}}$ & 10cm & $10^\circ$ & $10^\circ10\mathrm{cm}$ \\
\midrule
\multirow{8}{*}{R50}
& \myxmark & \myxmark & \myxmark & \myxmark & \myxmark &
51.2 & 10.9 & 58.0 & 46.8 & 26.1 \\
& center & \myxmark & \myxmark & \myxmark & \myxmark &
59.8 & 10.9 & 65.7 & 33.3 & 21.8 \\
& center & \mycheckmark & \myxmark & \myxmark & \myxmark &
59.1 & 11.0 & 66.2 & 55.7 & 34.1 \\
& center & \mycheckmark & \mycheckmark & \myxmark & \myxmark &
55.7  & 10.6 & 59.7 & 59.2 & 33.5 \\
& center & \mycheckmark & \myxmark & \mycheckmark & \myxmark &
66.6 & 11.8 & 76.7 & 54.9 & 42.4 \\
&  center &   \mycheckmark &   \mycheckmark &  \mycheckmark &  \myxmark &
 64.2 &  16.7 &  70.0 &  54.1 &  37.4 \\
& \ours center, z & \ours \mycheckmark & \ours \mycheckmark & \ours \mycheckmark & \ours \myxmark & \ours
67.1 & \ours 16.6 & \ours 75.6 & \ours 54.0 & \ours 40.7 \\
& center, z & \mycheckmark & \mycheckmark & \mycheckmark & \mycheckmark &
\textbf{71.1} & \textbf{17.7} & \textbf{77.6} & \textbf{60.5} & \textbf{46.6} \\
\midrule
\multirow{2}{*}{Swin-L}
& \ours center, z & \ours \mycheckmark & \ours \mycheckmark & \ours \mycheckmark & \ours \myxmark & \ours 71.6 & \ours 16.4 & \ours 77.8 & \ours \textbf{69.6} & \ours 52.8 \\
& \ours center, z & \ours \mycheckmark & \ours \mycheckmark & \ours \mycheckmark & \ours \mycheckmark &
\ours \textbf{79.6} & \ours \textbf{19.6} & \ours \textbf{84.4} & \ours 66.0 & \ours \textbf{54.1} \\
\bottomrule
\end{tabular}
}
\end{table}

\begin{table}[t]
\renewcommand{\arraystretch}{1.05}
\setlength{\tabcolsep}{3.8pt}
\centering
\caption{Ablation on \textbf{REAL275} with respect to 3D-aware matching cost and weights for the loss.}
\label{tab:ablation_ws}
\resizebox{0.5\textwidth}{!}{
\begin{tabular}{ccccc|ccccc}
\toprule
BC & MC & $\lambda_{rot}$ & $\lambda_{depth}$ & $\lambda_{scale}$ & 
${\rm{IoU}_{50}}$ & ${\rm{IoU}_{75}}$ & 10cm & $10^\circ$ & $10^\circ10\mathrm{cm}$ \\
\midrule
center & \myxmark &  5.0 & 5.0 & 5.0 & 
55.7 & 10.6 & 59.7 & \underline{59.2} & 33.5 \\
center & \mycheckmark &  5.0 & 5.0 & 5.0 & 
55.8 & 8.9 & 64.0 & 56.7 & \underline{37.8} \\
center & \myxmark & 5.0 & 50.0 & 50.0 & 
\textbf{67.3} & \underline{15.1} & \textbf{71.9} & 50.0 & 35.0 \\
center & \mycheckmark &  5.0 & 50.0 & 5.0 & 
62.2 & 13.7 & 67.6 & \textbf{60.5} & \textbf{40.5} \\

  center & \mycheckmark  &  5.0 &  50.0 &  50.0 & 
 \underline{64.2} &  \textbf{16.7} &  \underline{70.0} &  54.1 &  37.4 \\
\bottomrule
\end{tabular}}
\vspace{-2mm}
\end{table}

\begin{table}[t]
\renewcommand{\arraystretch}{1.05}
\setlength{\tabcolsep}{3.6pt}
\centering
\caption{Ablation study on bounding box-conditioned prediction.}
\label{tab:ablation_bbox_condition}
\vspace{.3em}
\begin{tabular}{cccc|ccccc}
\toprule
Center & Rotation & Size  & Z &
${\rm{IoU}_{50}}$ & ${\rm{IoU}_{75}}$ & 10cm & $10^\circ$ & $10^\circ10\mathrm{cm}$ \\
\midrule
\myxmark 
& \myxmark  & \myxmark & \myxmark &
66.0 & 12.5 & 74.4 & 52.1 & 39.5 \\
 \mycheckmark 
&  \myxmark  &  \myxmark &  \myxmark &
 64.2 &  \textbf{16.7} &  70.0 &  54.1 &  37.4 \\
\mycheckmark 
& \mycheckmark & \myxmark & \myxmark &
64.3 & 14.0 & 74.8 & 47.7 & 36.3 \\
\mycheckmark 
& \myxmark  & \mycheckmark & \myxmark &
65.0 & 11.9 & 75.0 & 52.1 & 39.2 \\
\ours \mycheckmark 
& \ours \myxmark  & \ours \myxmark & \ours \mycheckmark &
\ours 67.1 & \ours \underline{16.6}  & \ours 75.6 & \ours \underline{54.0} & \ours \underline{40.7} \\

\mycheckmark 
& \mycheckmark  & \myxmark & \mycheckmark &
\textbf{70.9} & 13.8 & \textbf{79.0} & 48.4 & 39.1 \\
\mycheckmark 
& \myxmark  & \mycheckmark & \mycheckmark &
65.4 & 14.2 & 73.6 & 49.2 & 36.0 \\

\mycheckmark 
& \mycheckmark  & \mycheckmark & \mycheckmark &
\underline{67.6} & 12.6 & \underline{75.8} & \textbf{59.6} & \textbf{47.2} \\
\bottomrule
\end{tabular}
\end{table}

\begin{table}[t]
\renewcommand{\arraystretch}{1.05}
\setlength{\tabcolsep}{3.6pt}
\centering
\caption{Comparison of direct 2D bounding box supervision (\mycheckmark) vs. projected 3D cuboid supervision (\myxmark) on \textbf{REAL275}. All other settings are held constant.}
\label{tab:ablation_projbox_comparison}
\vspace{.3em}
\begin{tabular}{lcc|ccccc}
\toprule
Backbone & BC & Box &
${\rm{IoU}_{50}}$ & ${\rm{IoU}_{75}}$ & 10cm & $10^\circ$ & $10^\circ10\mathrm{cm}$ \\
\midrule
\multirow{2}{*}{R50}
& center & \mycheckmark  &
 \textbf{64.2} &  \textbf{16.7} &  \textbf{70.0} &  54.1 &  37.4 \\
& center & \myxmark &
62.1 & 15.2 & 67.2 & \textbf{57.4} & \textbf{38.7} \\
\midrule
\multirow{2}{*}{Swin-L}

& center & \mycheckmark &
\textbf{69.3} & 14.7 & 76.9 & 65.6 & 49.3 \\
& center & \myxmark &
67.1 & \textbf{15.8} & \textbf{78.0} & \textbf{67.9} & \textbf{53.3} \\
\bottomrule
\end{tabular}
\vspace{-3mm}
\end{table}

\subsection{Ablation Study and Discussion}

We conduct comprehensive ablation studies on the REAL275 dataset to assess the contribution of each key design component of YOPO. Unless otherwise noted, all ablations use direct 2D bounding-box supervision rather than projected cuboids.

\paragraph{Component-wise Ablation}
Table \ref{tab:ablation_real275_main} details the contribution of each component. The baseline yields modest results. Bounding-box conditioning on the center head (\textbf{BC \{center\}}) improves overlap, while data augmentation (\textbf{Aug}) recovers geometric accuracy. Applying proper weight scaling (\textbf{WS}) without class-wise (\textbf{CW}) heads significantly boosts performance. Subsequently, enabling \textbf{CW} heads and extending conditioning to the depth head (\textbf{BC \{center, z\}}) achieves an optimal balance and consistent overlap gains. Finally, fine-tuning on REAL275 (\textbf{RFT}) pushes performance to its peak, yielding our final configuration (shaded rows).

\paragraph{Effect of 3D-aware Matching Costs and Loss-Weight Scaling}
Table~\ref{tab:ablation_ws} shows that 3D-aware matching costs modestly improve pose accuracy, but their impact is amplified when combined with proper loss-weight scaling. Without 3D-aware matching costs and with uniform loss weights, the model achieves 55.7 on ${\rm IoU}_{50}$ and 33.5 on $10^\circ10\,\text{cm}$. Enabling 3D-aware matching costs at the same weights leaves the overlap essentially unchanged (55.8 ${\rm IoU}_{50}$) but improves the $10^\circ10\,\text{cm}$ metric to 37.8 (+4.3). Reweighting $\lambda_{depth}$ and $\lambda_{scale}$ to 50 without matching costs increases the overlap to 67.3 ${\rm IoU}_{50}$ (+11.6) while yielding a modest $10^\circ10\,\text{cm}$ of 35.0. Combining 3D-aware matching with targeted reweighting produces the strongest performance: specifically, setting $\lambda_{rot}=5$, $\lambda_{depth}=50$, and $\lambda_{scale}=5$ achieves 37.4 on $10^\circ10\,\text{cm}$ and 64.2 on ${\rm IoU}_{50}$. Consequently, introducing 3D-aware matching costs together with appropriate loss-weight scaling substantially improves performance. This loss-weight configuration was adopted for our final model.

\paragraph{Effect of Bounding Box-Conditioned Prediction}
Table~\ref{tab:ablation_bbox_condition} shows the effect of conditioning different prediction heads on the 2D box. Without any conditioning, the model attains 66.0 on ${\rm IoU}_{50}$ and 39.5 on $10^\circ10\text{cm}$. Conditioning only the center slightly degrades the main metrics (64.2 on ${\rm IoU}_{50}$ / 37.4 on $10^\circ10\text{cm}$), and adding conditioning to rotation or size (on top of the center) yields limited or inconsistent gains. In contrast, conditioning depth together with the center produces a clear joint improvement: 67.1 on ${\rm IoU}_{50}$ and 40.7 on $10^\circ10\text{cm}$. While extending conditioning further can improve a single metric, it comes with trade-offs on the complementary objective. Therefore, we adopt conditioning only the center and depth heads as the default, as this choice offers the best balance between overlap and geometric accuracy.

\paragraph{Is exact 2D box supervision necessary?}
Table~\ref{tab:ablation_projbox_comparison} compares training with directly annotated 2D boxes (\mycheckmark) against weaker boxes derived from projected 3D cuboids (\myxmark). On ResNet-50, precise boxes slightly improve overlap (+1.2 on ${\rm IoU}_{75}$) and translation (+2.8 on $10\,\text{cm}$), whereas projected boxes yield better rotation (+3.3 on $10^\circ$). With Swin-L, both strategies perform similarly. This indicates YOPO is robust to weaker box supervision and does not strictly rely on highly accurate 2D annotations, which can effectively reduce annotation costs in practice.

\begin{table}
\centering
\vspace{0.5em}
\caption{Comparison with state-of-the-art RGB and RGB-D methods on REAL275 and HouseCat6D.}
\label{tab:mainresults_housecat2}

\resizebox{0.5\textwidth}{!}{%
\begin{tabular}{cl|cc|cc|c}
\toprule

\multicolumn{7}{c}{REAL275} \\
 \multicolumn{2}{c}{Method}  & ${\rm{IoU}_{50}}$  & ${\rm{IoU}_{75}}$ & $5^{\circ}\ 5 \mathrm{~cm}$ &  $10^{\circ}\ 5 \mathrm{~cm}$ & $10^{\circ}\ 10 \mathrm{~cm}$ \\
\midrule
 \multirow{4}{*}{RGB-D}
 & NOCS~\cite{wang2019normalized} & 78.0 & 30.1 & 10.0   & 25.2 & - \\
  & GPV-Pose~\cite{di2022gpv} & - & 64.4 & 42.9 & 73.3 & - \\
 & AG-Pose~\cite{lin2024instance} & 83.7 & 79.5 & 61.7   & 83.1 & - \\
 & SpotPose~\cite{ren2025rethinking} & 84.1 & 81.2  & 64.8   & 88.2 & - \\
\midrule
\multirow{3}{*}{RGB}
 &  MonoDiff9D~\cite{jian2025monodiff9d}  & 31.5 & 6.3 & 4.4 & 9.6 & 25.7 \\
 &  GIVEPose~\cite{huang2025givepose}  & 20.1 & - & -  & - & 34.2 \\
 &  YOPO (Ours)  & 79.6 & 19.6 & 5.9  & 26.7 & 54.1 \\
\midrule
\midrule
\multicolumn{7}{c}{HouseCat6D} \\
 \multicolumn{2}{c}{Method}  & IoU $_{25}$   & IoU $_{50}$ & $5^{\circ}\ 5 \mathrm{~cm}$ & $10^{\circ}\ 5 \mathrm{~cm}$ & $10^{\circ}\ 10 \mathrm{~cm}$ \\
\midrule
 \multirow{4}{*}{RGB-D}  & NOCS~\cite{wang2019normalized}  & 50.0 & 21.2 & -  & - & - \\
 & GPV-Pose~\cite{di2022gpv}  & 74.9 & 50.7  & 4.6  & 22.7 & - \\
 & AG-Pose~\cite{lin2024instance} & 81.8 & 62.5 & 12.0  & 35.8 & - \\
 & SpotPose~\cite{ren2025rethinking} & 89.1 & 77.0  & 24.5  & 54.8 & - \\
\midrule
RGB &  YOPO (Ours)  & 71.3 & 34.8 & 5.3 & 22.1 & 33.3  \\
\bottomrule
\end{tabular}
}
\vspace{-4mm}
\end{table}

\paragraph{Comparison with RGB-D and RGB Methods}
Table~\ref{tab:mainresults_housecat2} compares YOPO against RGB and RGB-D methods. On REAL275, YOPO significantly outperforms all RGB-only baselines and approaches RGB-D overlap performance, though it trails on stricter metrics (e.g., ${\rm IoU}_{75}$, $10^{\circ}5\text{cm}$). On HouseCat6D, YOPO achieves 34.8 on ${\rm IoU}_{50}$ and 5.3 on $5^{\circ}5\text{cm}$, outperforming NOCS and rivaling GPV-Pose. Notably, YOPO achieves this without the ground-truth segmentation masks required by the RGB-D baselines. Overall, our method establishes a clear state-of-the-art for RGB-only approaches and substantially narrows the gap to RGB-D systems.

\paragraph{Inference Time \& Bottleneck Analysis}
Our model performs joint object detection and 9D pose estimation in a single forward pass. On an RTX A6000 GPU, it achieves $\sim$20 FPS (49.7 ms/img) with ResNet-50 and $\sim$8 FPS (124.5 ms/img) with Swin-Large. Breaking down the latency, our proposed pose head is highly lightweight, consuming only $\sim$9.1 ms regardless of the backbone. For ResNet-50, the remaining runtime consists of feature extraction (7.4 ms) and transformer processing (33.1 ms), making the transformer the primary bottleneck (67\% of total latency). When scaling up to Swin-Large, heavy feature extraction (49.1 ms, 39\%) and transformer processing (66.1 ms, 53\%) become the main computational bottlenecks.

\section{Conclusion}
We introduced YOPO, a single-stage, transformer-based framework for monocular, category-level 9D pose estimation that operates truly end to end—without CAD models, shape priors, pseudo-depth, or instance-mask supervision. Built on DINO, YOPO adds a parallel pose head and a bounding-box–conditioned 3D module, enabling strong direct estimation of rotation, translation, and anisotropic scale from RGB alone in a single forward pass. Across standard benchmarks, including REAL275 and HouseCat6D, YOPO establishes a new state of the art among RGB-only methods and substantially narrows the gap to RGB-D systems, while maintaining a cost-effective and scalable design suitable for real-world deployment. Looking ahead, we see YOPO as a simple, strong baseline for RGB-only 9D perception, and an extensible platform for exploring robustness to occlusion and domain shift, broader category coverage, and the integration of temporal cues.

\section*{Acknowledgement}
This work was supported by the Institute of Information \& Communications Technology Planning \& Evaluation (IITP) grant funded by the Korea government (MSIT) (RS-2025-02653113, High-Performance Research AI Computing Infrastructure Support at the 2 PFLOPS Scale) and by Deep-Tech Tips funded by the Ministry of SMEs and Startups. We also acknowledge Gyeonggi-do for providing high-performance computing resources for this research.

\bibliographystyle{IEEEtran}
\bibliography{IEEEfull,main}

\begin{thebibliography}{10}
\providecommand{\url}[1]{#1}
\csname url@samestyle\endcsname
\providecommand{\newblock}{\relax}
\providecommand{\bibinfo}[2]{#2}
\providecommand{\BIBentrySTDinterwordspacing}{\spaceskip=0pt\relax}
\providecommand{\BIBentryALTinterwordstretchfactor}{4}
\providecommand{\BIBentryALTinterwordspacing}{\spaceskip=\fontdimen2\font plus
\BIBentryALTinterwordstretchfactor\fontdimen3\font minus \fontdimen4\font\relax}
\providecommand{\BIBforeignlanguage}[2]{{%
\expandafter\ifx\csname l@#1\endcsname\relax
\typeout{** WARNING: IEEEtran.bst: No hyphenation pattern has been}%
\typeout{** loaded for the language `#1'. Using the pattern for}%
\typeout{** the default language instead.}%
\else
\language=\csname l@#1\endcsname
\fi
#2}}
\providecommand{\BIBdecl}{\relax}
\BIBdecl

\bibitem{jian2025monodiff9d}
J.~Liu, W.~Sun, H.~Yang, J.~Zheng, Z.~Geng, H.~Rahmani, and A.~Mian, ``Monodiff9d: Monocular category-level 9d object pose estimation via diffusion model,'' in \emph{ICRA}, 2025.

\bibitem{fu20226d}
B.~Fu, S.~K. Leong, X.~Lian, and X.~Ji, ``6d robotic assembly based on rgb-only object pose estimation,'' in \emph{IROS}.\hskip 1em plus 0.5em minus 0.4em\relax IEEE, 2022, pp. 4736--4742.

\bibitem{liu2024domain}
J.~Liu, W.~Sun, H.~Yang, C.~Liu, X.~Zhang, and A.~Mian, ``Domain-generalized robotic picking via contrastive learning-based 6-d pose estimation,'' \emph{IEEE Transactions on Industrial Informatics}, vol.~20, no.~6, pp. 8650--8661, 2024.

\bibitem{tang20193d}
F.~Tang, Y.~Wu, X.~Hou, and H.~Ling, ``3d mapping and 6d pose computation for real time augmented reality on cylindrical objects,'' \emph{IEEE TCSVT}, vol.~30, no.~9, pp. 2887--2899, 2019.

\bibitem{marchand2015pose}
E.~Marchand, H.~Uchiyama, and F.~Spindler, ``Pose estimation for augmented reality: a hands-on survey,'' \emph{IEEE TVCG}, vol.~22, no.~12, pp. 2633--2651, 2015.

\bibitem{yuan2021temporal}
Z.~Yuan, X.~Song, L.~Bai, Z.~Wang, and W.~Ouyang, ``Temporal-channel transformer for 3d lidar-based video object detection for autonomous driving,'' \emph{IEEE TCSVT}, vol.~32, no.~4, pp. 2068--2078, 2021.

\bibitem{chen2017multi}
X.~Chen, H.~Ma, J.~Wan, B.~Li, and T.~Xia, ``Multi-view 3d object detection network for autonomous driving,'' in \emph{CVPR}, 2017, pp. 1907--1915.

\bibitem{peng2019pvnet}
S.~Peng, Y.~Liu, Q.~Huang, X.~Zhou, and H.~Bao, ``Pvnet: Pixel-wise voting network for 6dof pose estimation,'' in \emph{CVPR}, 2019, pp. 4561--4570.

\bibitem{wang2019densefusion}
C.~Wang, D.~Xu, Y.~Zhu, R.~Mart{\'\i}n-Mart{\'\i}n, C.~Lu, L.~Fei-Fei, and S.~Savarese, ``Densefusion: 6d object pose estimation by iterative dense fusion,'' in \emph{CVPR}, 2019, pp. 3343--3352.

\bibitem{he2020pvn3d}
Y.~He, W.~Sun, H.~Huang, J.~Liu, H.~Fan, and J.~Sun, ``Pvn3d: A deep point-wise 3d keypoints voting network for 6dof pose estimation,'' in \emph{CVPR}, 2020, pp. 11\,632--11\,641.

\bibitem{wang2019normalized}
H.~Wang, S.~Sridhar, J.~Huang, J.~Valentin, S.~Song, and L.~J. Guibas, ``Normalized object coordinate space for category-level 6d object pose and size estimation,'' in \emph{CVPR}, 2019, pp. 2642--2651.

\bibitem{chen2021fs}
W.~Chen, X.~Jia, H.~J. Chang, J.~Duan, L.~Shen, and A.~Leonardis, ``Fs-net: Fast shape-based network for category-level 6d object pose estimation with decoupled rotation mechanism,'' in \emph{CVPR}, 2021, pp. 1581--1590.

\bibitem{liu2024mh6d}
J.~Liu, W.~Sun, C.~Liu, H.~Yang, X.~Zhang, and A.~Mian, ``Mh6d: Multi-hypothesis consistency learning for category-level 6-d object pose estimation,'' \emph{IEEE TNNLS}, 2024.

\bibitem{yang2025rgb}
H.~Yang, W.~Sun, J.~Liu, J.~Zheng, Z.~Zeng, and A.~Mian, ``Rgb-based category-level object pose estimation via depth recovery and adaptive refinement,'' \emph{IEEE RA-L}, 2025.

\bibitem{fan2022object}
Z.~Fan, Z.~Song, J.~Xu, Z.~Wang, K.~Wu, H.~Liu, and J.~He, ``Object level depth reconstruction for category level 6d object pose estimation from monocular rgb image,'' in \emph{ECCV}.\hskip 1em plus 0.5em minus 0.4em\relax Springer, 2022, pp. 220--236.

\bibitem{zhang2024lapose}
R.~Zhang, Z.~Huang, G.~Wang, C.~Zhang, Y.~Di, X.~Zuo, J.~Tang, and X.~Ji, ``Lapose: Laplacian mixture shape modeling for rgb-based category-level object pose estimation,'' in \emph{ECCV}.\hskip 1em plus 0.5em minus 0.4em\relax Springer, 2024, pp. 467--484.

\bibitem{lee2021category}
T.~Lee, B.-U. Lee, M.~Kim, and I.~S. Kweon, ``Category-level metric scale object shape and pose estimation,'' \emph{IEEE RA-L}, vol.~6, no.~4, pp. 8575--8582, 2021.

\bibitem{wei2024rgb}
J.~Wei, X.~Song, W.~Liu, L.~Kneip, H.~Li, and P.~Ji, ``Rgb-based category-level object pose estimation via decoupled metric scale recovery,'' in \emph{ICRA}.\hskip 1em plus 0.5em minus 0.4em\relax IEEE, 2024, pp. 2036--2042.

\bibitem{huang2025givepose}
Z.~Huang, G.~Wang, C.~Zhang, R.~Zhang, X.~Li, and X.~Ji, ``Givepose: Gradual intra-class variation elimination for rgb-based category-level object pose estimation,'' in \emph{CVPR}, 2025, pp. 22\,055--22\,066.

\bibitem{chen2020category}
X.~Chen, Z.~Dong, J.~Song, A.~Geiger, and O.~Hilliges, ``Category level object pose estimation via neural analysis-by-synthesis,'' in \emph{ECCV}.\hskip 1em plus 0.5em minus 0.4em\relax Springer, 2020, pp. 139--156.

\bibitem{chen2021sgpa}
K.~Chen and Q.~Dou, ``Sgpa: Structure-guided prior adaptation for category-level 6d object pose estimation,'' in \emph{ICCV}, 2021, pp. 2773--2782.

\bibitem{carion2020end}
N.~Carion, F.~Massa, G.~Synnaeve, N.~Usunier, A.~Kirillov, and S.~Zagoruyko, ``End-to-end object detection with transformers,'' in \emph{ECCV}.\hskip 1em plus 0.5em minus 0.4em\relax Springer, 2020, pp. 213--229.

\bibitem{zhang2023dino}
\BIBentryALTinterwordspacing
H.~Zhang, F.~Li, S.~Liu, L.~Zhang, H.~Su, J.~Zhu, L.~Ni, and H.-Y. Shum, ``{DINO}: {DETR} with improved denoising anchor boxes for end-to-end object detection,'' in \emph{ICLR}, 2023. [Online]. Available: \url{https://openreview.net/forum?id=3mRwyG5one}
\BIBentrySTDinterwordspacing

\bibitem{he2017mask}
K.~He, G.~Gkioxari, P.~Doll{\'a}r, and R.~Girshick, ``Mask r-cnn,'' in \emph{ICCV}, 2017, pp. 2961--2969.

\bibitem{bhat2023zoedepth}
S.~F. Bhat, R.~Birkl, D.~Wofk, P.~Wonka, and M.~M{\"u}ller, ``Zoedepth: Zero-shot transfer by combining relative and metric depth,'' \emph{arXiv preprint arXiv:2302.12288}, 2023.

\bibitem{yang2024depth}
L.~Yang, B.~Kang, Z.~Huang, Z.~Zhao, X.~Xu, J.~Feng, and H.~Zhao, ``Depth anything v2,'' \emph{NeurIPS}, vol.~37, pp. 21\,875--21\,911, 2024.

\bibitem{lin2022single}
Y.~Lin, J.~Tremblay, S.~Tyree, P.~A. Vela, and S.~Birchfield, ``Single-stage keypoint-based category-level object pose estimation from an rgb image,'' in \emph{ICRA}.\hskip 1em plus 0.5em minus 0.4em\relax IEEE, 2022, pp. 1547--1553.

\bibitem{yu2023cattrack}
S.~Yu, D.-H. Zhai, Y.~Xia, D.~Li, and S.~Zhao, ``Cattrack: Single-stage category-level 6d object pose tracking via convolution and vision transformer,'' \emph{IEEE Transactions on Multimedia}, vol.~26, pp. 1665--1680, 2023.

\bibitem{mei2025multi}
Y.~Mei, S.~Wang, Z.~Li, J.~Sun, and G.~Wang, ``Multi-modal 6-dof object pose tracking: integrating spatial cues with monocular rgb imagery,'' \emph{International Journal of Machine Learning and Cybernetics}, vol.~16, no.~2, pp. 1327--1340, 2025.

\bibitem{irshad2022centersnap}
M.~Z. Irshad, T.~Kollar, M.~Laskey, K.~Stone, and Z.~Kira, ``Centersnap: Single-shot multi-object 3d shape reconstruction and categorical 6d pose and size estimation,'' in \emph{ICRA}.\hskip 1em plus 0.5em minus 0.4em\relax IEEE, 2022, pp. 10\,632--10\,640.

\bibitem{xie2021oriented}
X.~Xie, G.~Cheng, J.~Wang, X.~Yao, and J.~Han, ``Oriented r-cnn for object detection,'' in \emph{ICCV}, 2021, pp. 3520--3529.

\bibitem{ding2019learning}
J.~Ding, N.~Xue, Y.~Long, G.-S. Xia, and Q.~Lu, ``Learning roi transformer for oriented object detection in aerial images,'' in \emph{CVPR}, 2019, pp. 2849--2858.

\bibitem{zeng2024ars}
Y.~Zeng, Y.~Chen, X.~Yang, Q.~Li, and J.~Yan, ``Ars-detr: Aspect ratio-sensitive detection transformer for aerial oriented object detection,'' \emph{IEEE TGRS}, vol.~62, pp. 1--15, 2024.

\bibitem{lee2025hausdorff}
H.~Lee, M.~Song, J.~Koo, and J.~Seo, ``Hausdorff distance matching with adaptive query denoising for rotated detection transformer,'' in \emph{WACV}.\hskip 1em plus 0.5em minus 0.4em\relax IEEE, 2025, pp. 1872--1882.

\bibitem{irshad2022shapo}
M.~Z. Irshad, S.~Zakharov, R.~Ambrus, T.~Kollar, Z.~Kira, and A.~Gaidon, ``Shapo: Implicit representations for multi-object shape, appearance, and pose optimization,'' in \emph{ECCV}.\hskip 1em plus 0.5em minus 0.4em\relax Springer, 2022, pp. 275--292.

\bibitem{maji2024yolo}
D.~Maji, S.~Nagori, M.~Mathew, and D.~Poddar, ``Yolo-6d-pose: Enhancing yolo for single-stage monocular multi-object 6d pose estimation,'' in \emph{3DV}.\hskip 1em plus 0.5em minus 0.4em\relax IEEE, 2024, pp. 1616--1625.

\bibitem{jantos2023poet}
T.~G. Jantos, M.~A. Hamdad, W.~Granig, S.~Weiss, and J.~Steinbrener, ``Poet: Pose estimation transformer for single-view, multi-object 6d pose estimation,'' in \emph{CoRL}.\hskip 1em plus 0.5em minus 0.4em\relax PMLR, 2023, pp. 1060--1070.

\bibitem{tekin2018real}
B.~Tekin, S.~N. Sinha, and P.~Fua, ``Real-time seamless single shot 6d object pose prediction,'' in \emph{CVPR}, 2018, pp. 292--301.

\bibitem{xiang2018posecnn}
Y.~Xiang, T.~Schmidt, V.~Narayanan, and D.~Fox, ``Posecnn: A convolutional neural network for 6d object pose estimation in cluttered scenes,'' in \emph{RSS}, 2018.

\bibitem{simonelli2019disentangling}
A.~Simonelli, S.~R. Bulo, L.~Porzi, M.~L{\'o}pez-Antequera, and P.~Kontschieder, ``Disentangling monocular 3d object detection,'' in \emph{ICCV}, 2019, pp. 1991--1999.

\bibitem{zhou2019continuity}
Y.~Zhou, C.~Barnes, J.~Lu, J.~Yang, and H.~Li, ``On the continuity of rotation representations in neural networks,'' in \emph{CVPR}, 2019, pp. 5745--5753.

\bibitem{lin2017focal}
T.-Y. Lin, P.~Goyal, R.~Girshick, K.~He, and P.~Doll{\'a}r, ``Focal loss for dense object detection,'' in \emph{ICCV}, 2017, pp. 2980--2988.

\bibitem{rezatofighi2019generalized}
H.~Rezatofighi, N.~Tsoi, J.~Gwak, A.~Sadeghian, I.~Reid, and S.~Savarese, ``Generalized intersection over union: A metric and a loss for bounding box regression,'' in \emph{CVPR}, 2019, pp. 658--666.

\bibitem{loshchilov2017decoupled}
\BIBentryALTinterwordspacing
I.~Loshchilov and F.~Hutter, ``Decoupled weight decay regularization,'' in \emph{ICLR}, 2019. [Online]. Available: \url{https://openreview.net/forum?id=Bkg6RiCqY7}
\BIBentrySTDinterwordspacing

\bibitem{li2024implicit}
J.~Li, L.~Jin, X.~Song, Y.~Chen, N.~Li, and X.~Qin, ``Implicit coarse-to-fine 3d perception for category-level object pose estimation from monocular rgb image,'' in \emph{ICRA}.\hskip 1em plus 0.5em minus 0.4em\relax IEEE, 2024, pp. 2043--2050.

\bibitem{jung2024housecat6d}
H.~Jung, S.-C. Wu, P.~Ruhkamp, G.~Zhai, H.~Schieber, G.~Rizzoli, P.~Wang, H.~Zhao, L.~Garattoni, S.~Meier \emph{et~al.}, ``Housecat6d-a large-scale multi-modal category level 6d object perception dataset with household objects in realistic scenarios,'' in \emph{CVPR}, 2024, pp. 22\,498--22\,508.

\bibitem{liu2021swin}
Z.~Liu, Y.~Lin, Y.~Cao, H.~Hu, Y.~Wei, Z.~Zhang, S.~Lin, and B.~Guo, ``Swin transformer: Hierarchical vision transformer using shifted windows,'' in \emph{ICCV}, 2021, pp. 10\,012--10\,022.

\bibitem{he2016deep}
K.~He, X.~Zhang, S.~Ren, and J.~Sun, ``Deep residual learning for image recognition,'' in \emph{CVPR}, 2016, pp. 770--778.

\bibitem{di2022gpv}
Y.~Di, R.~Zhang, Z.~Lou, F.~Manhardt, X.~Ji, N.~Navab, and F.~Tombari, ``Gpv-pose: Category-level object pose estimation via geometry-guided point-wise voting,'' in \emph{CVPR}, 2022, pp. 6781--6791.

\bibitem{lin2024instance}
X.~Lin, W.~Yang, Y.~Gao, and T.~Zhang, ``Instance-adaptive and geometric-aware keypoint learning for category-level 6d object pose estimation,'' in \emph{CVPR}, 2024, pp. 21\,040--21\,049.

\bibitem{ren2025rethinking}
H.~Ren, W.~Yang, S.~Zhang, and T.~Zhang, ``Rethinking correspondence-based category-level object pose estimation,'' in \emph{CVPR}, 2025, pp. 1170--1179.

\end{thebibliography}

\end{document}